\documentclass{article}

\usepackage{arxiv}

\usepackage[utf8]{inputenc} 
\usepackage[T1]{fontenc}    
\usepackage{hyperref}       
\usepackage{url}            
\usepackage{booktabs}       
\usepackage{amsfonts}       
\usepackage{microtype}      

\usepackage{amsmath}

\usepackage{graphicx}

\usepackage{systeme}
\usepackage{bm} 			
\usepackage{cite}

\usepackage{amsthm}			

\theoremstyle{definition}
\newtheorem{definition}{Definition}

\title{Stratified cross-validation for unbiased and privacy-preserving federated learning}

\author{
Romain Bey \\
Université de Paris \\
CRESS, INSERM, INRA \\
F-75004 Paris, France
\And
Romain Goussault \\
Service Oncodermatologie \\
CHU Nantes, CIC 1413, CRCINA Inserm 1232 \\
Nantes, France
\And
Mehdi Benchoufi \\
Université de Paris \\
CRESS, INSERM, INRA \\
F-75004 Paris, France
\And
Rapha\"el Porcher\\
Université de Paris \\
CRESS, INSERM, INRA \\
F-75004 Paris, France \\ \texttt{raphael.porcher@aphp.fr}}

\begin{document}
\maketitle

\begin{abstract}
\subsection*{Objective}
We introduce \textit{stratified cross-validation}, a validation methodology that is compatible with privacy-preserving federated learning and that prevents data leakage caused by duplicates of electronic health records (EHR).

\subsection*{Materials and methods}
\textit{Stratified cross-validation} complements cross-validation with an initial stratification of EHR in folds containing similar patients, thus ensuring that duplicates of a record are jointly present either in training or in validation folds. Monte Carlo simulations are performed to investigate the properties of \textit{stratified cross-validation} in the case of a model data analysis.

\subsection*{Results}
In situations where duplicated EHR could induce over-optimistic estimations of accuracy, applying \textit{stratified cross-validation} prevented this bias, while not requiring full deduplication. However, a pessimistic bias might appear if the covariate used for the stratification was strongly associated with the outcome.

\subsection*{Discussion}
Although \textit{stratified cross-validation} presents low computational overhead, to be efficient it requires the preliminary identification of a covariate that is both shared by duplicated records and weakly associated with the outcome. When available, the hash of a personal identifier or a patient's date of birth provides such a covariate. On the contrary, pseudonymization interferes with \textit{stratificatied cross-validation} as it may break the equality of the stratifying covariate among duplicates.

\subsection*{Conclusion}
\textit{Stratified cross-validation} is an easy-to-implement methodology that prevents data leakage when a model is trained on distributed EHR that contain duplicates, while preserving privacy.

\end{abstract}


\section*{Background and significance}

The large-scale collection of data and its analysis by artificial intelligence (AI) algorithms have led to new scientific discoveries and huge expectations for the near future \cite{esteva_dermatologist-level_2017, hosny_artificial_2018, komorowski_artificial_2018, rajkomar_scalable_2018, rahimian_predicting_2018}. Although AI algorithms (random forest, gradient boosting, neural networks, etc.\cite{hastie_elements_2009}) provide powerful tools, they are difficult to develop as they generally require large training datasets to reach reasonable performances \cite{van_der_ploeg_modern_2014} and often detailed, high dimensional records about individuals. Beyond the technical challenges that such a data collection represents, storing and analyzing a large amount of personally identifying information (PII) may moreover imply serious risks regarding privacy, and recent research projects have been hindered by public opinion concerns \cite{powles_google_2017,caldicott_review_2016}. These concerns are not unfounded as re-identification attacks have regularly broken the anonymity of large datasets \cite{homer_resolving_2008, bohannon_genetics._2013, gymrek_identifying_2013, rocher_estimating_2019}. To address these risks new regulations that impose higher security standards have been introduced \cite{price_privacy_2019}. Technically, it moreover appears necessary to complement classical anonymization techniques as they are intrinsically limited in the case of high dimensional data \cite{aggarwal_k-anonymity_2005, brickell_cost_2008}. Handling and analyzing securely such data therefore requires moving from the \textit{anonymize, release and forget} approach to configurations where a data curator \textit{secures and controls} the use of data that remain to some extent identifying \cite{de_montjoye_privacy-conscientious_2018}. In the latter configuration, the question arises as to which organization should play the role of the trusted data curator. In the case of medical records, patients and IT managers have been reluctant to devote this role to centralized private or public organisations \cite{powles_google_2017,caldicott_review_2016,vest_hospitals_2018}, limiting large-scale research on electronic health records (EHR). To solve this issue a technique called federated learning has recently been proposed. This technique enables the training of AI models while keeping records in decentralized trusted data warehouses curated for instance by hospitals \cite{wu_grid_2012,lu_webdisco:_2015,shokri_privacy-preserving_2015, mcmahan_communication-efficient_2017, bonawitz_practical_2017,kairouz_advances_2019,bonawitz_towards_2019}. Federated learning appears as a promising privacy-enhancing technique that avoids single points of failure, and it is  currently being developed and tested in various projects worldwide \cite{raisaro_addressing_2017, raisaro_medco:_2019, ryffel_generic_2018, galtier_substra:_2019, duan_learning_2019}.

In addition to privacy concerns, recent controversies indicate that many AI models may have been validated improperly, shedding doubt on the performances that have been advertized \cite{lazer_big_2014, dressel_accuracy_2018, kiraly_nips_2018,park_methodologic_2018, vollmer_machine_2018}. One of the most frequent sources of bias in performance estimation is the data leakage that occurs when data used for validation and training are correlated \cite{kaufman_leakage_2012, harron_evaluating_2014, luo_guidelines_2016, saeb_need_2017}. Avoiding data leakage requires building training and validation datasets in such a way that all the data related to a given individual are exclusively in the former or the latter. This dataset building procedure may be compromised by the presence of different records related to the same individual: this risk is far from being negligible as it has been shown that up to 15\% of records in medical information systems are duplicates \cite{mccoy_matching_2013} and that many patients have records in multiple hospitals \cite{everson_gaps_2018}. Data leakage induced by duplicates may be especially important in the case of AI models, as they are often trained on large real-world datasets such as EHR, that have not been curated for research \cite{rajkomar_scalable_2018}. To address data leakage caused by duplicates, deduplication algorithms, often called record linkage algorithms, have been developed that rely on various deterministic or probabilistic methods \cite{harron_methodological_2015}.

Although federated learning and deduplication algorithms address privacy and validation issues respectively, they cannot be easily combined. Indeed, deduplication relies on the comparison of PII through the computation of a similarity index established between two potentially duplicated records, whereas federated learning avoids PII exchange and therefore prevents their comparison when records are located in different data warehouses. Consequently, detecting duplicates of a given record that are present in two different hospitals while preserving their privacy appears challenging. Some protocols have been developed and deployed to enable privacy-preserving deduplication in a federated learning setting, but they still require further research to be scalable \cite{vatsalan_taxonomy_2013, yigzaw_secure_2017, laud_privacy-preserving_2018}. In this paper, we propose an easier-to-implement approach that makes it possible to avoid data leakage in a federated learning setting while not relying on deduplication algorithms. We consider the classical cross-validation technique for performance estimation, and complement it by an initial stratification of datasets. Whereas some stratification techniques have already been combined with cross-validation in order to limit disparities between randomly chosen folds \cite{diamantidis_unsupervised_2000}, in this article we extend the use of stratification to avoid data leakage in the case of undetected data duplicates. 

This article is organised as follows. In \textit{Materials and Methods}, we describe the \textit{stratified cross-validation} methodology we propose and the synthetic datasets we study. In \textit{Results}, we simulate a data analysis in a federated learning setting following different validation strategies, the performances and limitations of which are detailed in \textit{Discussion} section.

\section*{Materials and methods}

\subsection*{Stratified cross-validation}

\begin{figure}
\centering
\includegraphics[width=12cm]{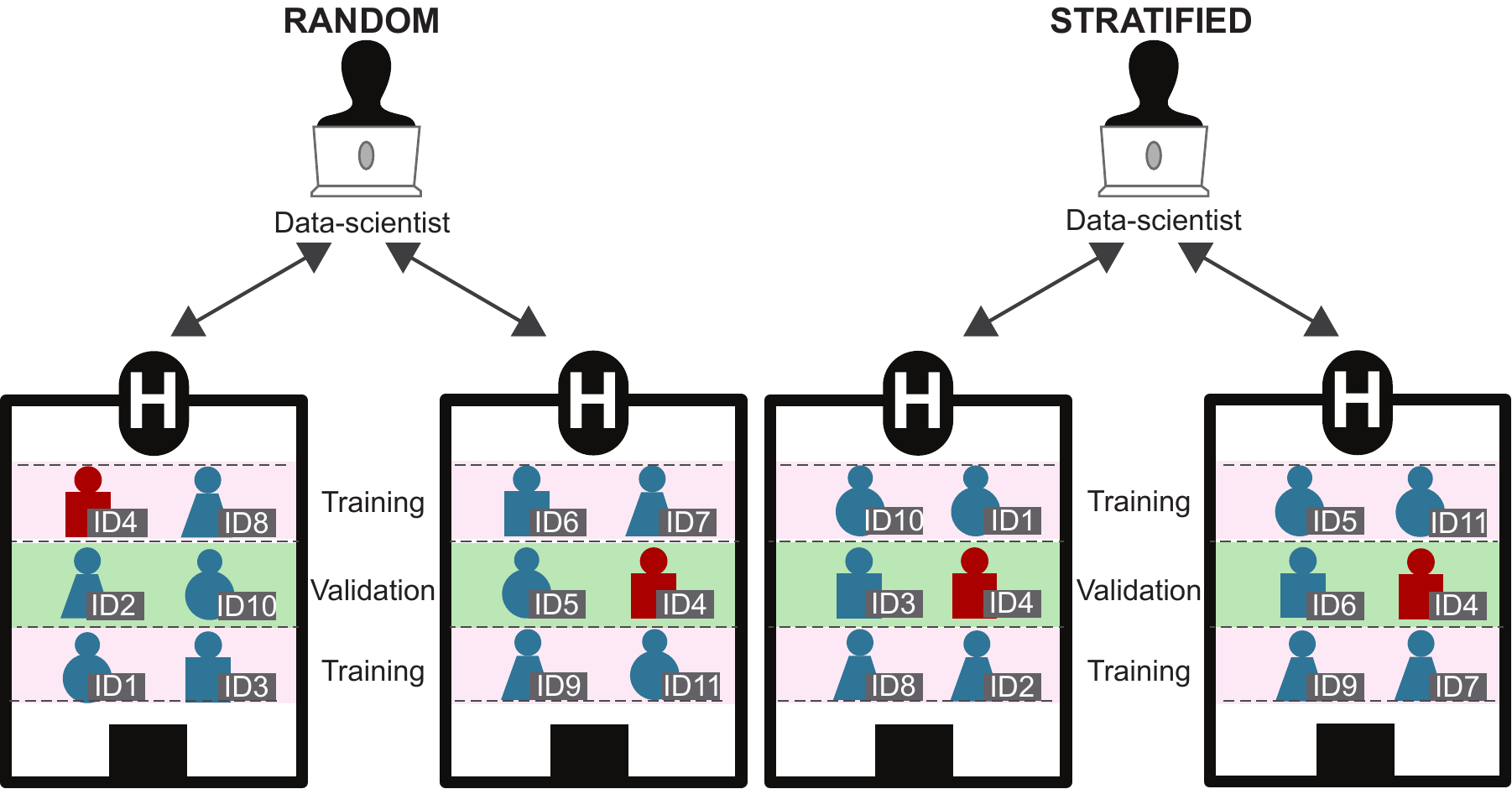}
\caption{Privacy-preserving federated learning: analysis by a data scientist of medical records (blue and red individuals) distributed in two hospitals without extracting personally identifying information (PII). One individual's record is duplicated in the two hospitals (red, ID4), due for instance to multiple admissions. The performances of a model are estimated through cross-validation, partitioning the datasets in training and validation folds either randomly (left) or through stratification, \textit{i.e.} grouping  similar patients in folds (right). Whereas duplicated records (red) may be simultaneously in training and validation folds when random partitioning is applied, thus causing data leakage, this risk is circumvented by stratification.}
\label{Fig1}
\end{figure}

We consider a model $f$ that computes a predicted $\tilde{y}$ of an outcome $y$ using covariates $\bm{x}=(x_1, x_2, ..., x_{m})$: $\tilde{y}=f(\bm{x})$. We consider a performance metric that we want to maximize and that is computed as the expectation of a function $h(y,\tilde{y})$. In the case of accuracy $h(y,\tilde{y}) = I(y=\tilde{y})$, where the function $I(A)$ equals 1 if $A$ is true and 0 otherwise. A record $\bm{r}=(\bm{x}, y)$ is a point in a mathematical space $\Omega$ that gathers for an individual her covariates $\bm{x}$ and her realized outcome $y$, and a dataset $\mathcal{D}$ is a collection of records. A dataset $\mathcal{D}$ contains duplicates when there are two records $\bm{r}_i$ and $\bm{r}_j$ with $i\neq j$ such that $\bm{r}_i=\bm{r}_j$. For the sake of simplicity we limit ourselves to exact duplicates but as discussed later our conclusions apply also to inaccurate duplicates, caused for instance by flawed or incomplete recording of data related to a given individual. We consider hereafter that there are never two individuals with perfectly identical records implying that equal records always relate to the same individual that has been registered more than once. We moreover consider that individuals whose records are duplicated are distributed uniformly in the population.

Cross-validation is a statistical technique commonly used to estimate the performances of a model \cite{hastie_elements_2009}. In cross-validation a dataset $\mathcal{D}$ is partitioned in $k$ folds (disjoint subsets) $\mathcal{D}_i$, with $i=1,..,k$. For each fold $\mathcal{D}_i$ the statistical model is trained on all the folds apart from fold $\mathcal{D}_i$ (training), and its performances are estimated on fold $\mathcal{D}_i$ (validation). The average of per-fold estimates is used as the estimate of the performances of the same model that would have been trained on all the records. Classical cross-validation often relies on a random partitioning of the dataset in $k$ folds. In that case duplicated records may be simultaneously present in training and validation folds thus inducing data leakage and yielding over-optimistic estimates of performance compared to cross-validation without duplicates, that we consider hereafter as the unbiased estimate. Although in that case data leakage is a consequence of duplicated records, it is possible to avoid it without complete deduplication. Indeed, data leakage occurs only when duplicated records are used both for training and validation. Ensuring that all duplicated records related to a given individual are used only for training or only for validation therefore prevents the risk of data leakage.

We consider training and validation of a model in a federated learning setting where records are distributed in different hospitals. In that case, many datasets related to different hospitals are jointly analysed by an external data scientist without exchanging PII (see Figure \ref{Fig1}). Cross-validation in a federated learning setting involves partitioning the dataset $\mathcal{D}^{(\alpha)}$ of each hospital $\alpha$ in $k$ folds $\mathcal{D}^{(\alpha)}_i$ with $i=1,..,k$. Folds are then merged over hospitals to obtain global folds: $\mathcal{D}_i=\cup_{\alpha}\mathcal{D}_i^{(\alpha)}$. To better characterize the presence of duplicates in a federated learning setting we introduce the following definitions:
\begin{definition}
\textbf{Intra-hospital deduplication}: For all records $\bm{r}_i, \bm{r}_j \in \mathcal{D}^{(\alpha)}$ in a hospital $\alpha$ with $i\neq j$ we have $\bm{r}_i \neq \bm{r}_j$.
\label{Def1}
\end{definition}
\begin{definition}
\textbf{Inter-hospital deduplication}: For all records $\bm{r}_i\in \mathcal{D}^{(\alpha)}$, $\bm{r}_j \in \mathcal{D}^{(\beta)}$ in different hospitals $\alpha\neq \beta$ we have $\bm{r}_i \neq \bm{r}_j$.
\label{Def2}
\end{definition}
Datasets $\mathcal{D}$ that jointly fulfill definitions \ref{Def1} and \ref{Def2} are completely deduplicated: \textit{i.e.} there are no two identical records $\bm{r}_i=\bm{r}_j$ with $i\neq j $ in the whole dataset. As explained above, although many deduplication techniques have been developed to fulfill definition \ref{Def1}, fulfilling definition \ref{Def2} remains challenging without loosing the privacy-enhancing advantage of federated learning. We therefore consider a weaker definition of deduplication that is sufficient to avoid data leakage between folds:
\begin{definition}
\textbf{Inter-fold deduplication}: For all records $\bm{r}_i\in \mathcal{D}^{(\alpha)}_m$, $\bm{r}_j\in \mathcal{D}^{(\beta)}_n$ related to different fold indexes $m\neq n$ we have $\bm{r}_i\neq \bm{r}_j$.
\label{Def3}
\end{definition}
Definition \ref{Def3} is weaker than definitions \ref{Def1} and \ref{Def2} as it can be fulfilled without removing all duplicates if one ensures instead that duplicates of a given record are present in the same fold. We propose hereafter a technique to create folds that fulfill definition \ref{Def3}. We consider a partition of the record space $\Omega$ in $k$ subspaces $\Omega_i$: $\Omega = \cup_{i=1}^{k} \Omega_i$ and $\Omega_i \cap \Omega_j = \emptyset$ if $i \neq j$. Such a partition can be realized stratifying $\Omega$ relatively to one covariate, and we refer to such a partition as a stratification. Once a stratification has been defined, each hospital $\alpha$ dataset $\mathcal{D}^{(\alpha)}$ can be partitioned in folds $i=1,2,...,k$ as follows:
\begin{equation}
  \left\{
      \begin{aligned}
        \mathcal{D}^{(\alpha)}_1 &=& \mathcal{D}^{(\alpha)}\cap \Omega_1\\
        &...&\\
        \mathcal{D}^{(\alpha)}_i &=& \mathcal{D}^{(\alpha)}\cap \Omega_i\\
        &...& \\
         \mathcal{D}^{(\alpha)}_k &=& \mathcal{D}^{(\alpha)}\cap \Omega_k
      \end{aligned}
    \right.
\label{Eq1}
\end{equation}
Partitioning each hospital dataset $\mathcal{D}^{(\alpha)}$ using a given stratification leads for $i\neq j$ to: $\mathcal{D}^{(\alpha)}_{i}\cap \mathcal{D}^{(\beta)}_j = (\mathcal{D}^{(\alpha)}\cap \Omega_i ) \cap (\mathcal{D}^{(\beta)} \cap \Omega_j) = (\mathcal{D}^{(\alpha)}\cap \mathcal{D}^{(\beta)}) \cap (\Omega_i \cap \Omega_j ) =\emptyset$ and the definition \ref{Def3} is therefore fulfilled. Combining stratification technique equation (\ref{Eq1}) with classical cross-validation constitutes the validation methodology that we call \textit{stratified cross-validation}.

Although \textit{stratified cross-validation} prevents over-optimistic estimations of performance induced by duplicates, it does not systematically provide an unbiased estimator compared to cross-validation in the absence of duplicates. A stratification procedure may indeed induce training and validation folds featuring different covariate distributions and covariate-outcome associations. A model trained and validated on such folds tends to overfit the training population and to be unfit for a generalization to the validation population, and  \textit{stratified cross-validation} appears akin to the external validation on a new population. Validating externally a model on a population coming from a different hospital is commonly recognized as a proof of quality as it measures the generalizability of a model to new care contexts, but \textit{stratified cross-validation} unfortunately does not measure this relevant inter-hospital generalizability as folds cannot be identified with hospitals. The stratification procedure should therefore be defined as to minimize the irrelevant pessimistic bias associated with the heterogeneity of folds populations. An ideal stratifying covariate would therefore be a covariate shared by duplicates but fully independent of the other covariates and of the outcome, as it would provide folds that would be statistically equivalent. Such a stratifying covariate is often not available as one only records covariates that are associated to some extent with patient's medical condition, and a challenge of \textit{stratified cross-validation} consists in finding a surrogate stratifying covariate that is weakly correlated to the other covariates and to the outcome. In the following section we run simulations to investigate and discuss the impact of various stratification strategies.

\begin{center}
\fbox{\begin{minipage}{30em}
\textit{Stratified cross-validation}
\begin{itemize}
\item Choose or create a stratifying covariate $x_{str}$ that is weakly associated with the other covariates and with the outcome.
\item Define thresholds $t_{0}$, $t_1$, ..., $t_k$ with $k$ the number of folds, such that there are approximately the same amount of records fulfilling $t_i<x_{str}<t_{i+1}$ for each $i$.
\item Associate each record with the fold index $i$ that fulfills $t_i<x_{str}<t_{i+1}$.
\item Group all records with the same fold index $i$ in inter-hospital folds $\mathcal{D}_i$ and apply cross-validation on these folds.
\end{itemize}
\end{minipage}}
\end{center}

\subsection*{Simulation}

In order to study the properties of \textit{stratified cross-validation} we simulate data analysis in a federated learning setting in presence of duplicates. We generate synthetic datasets in which a binary outcome $y$ depends on 10 covariates $x_1, x_2, ..., x_{10}$. Covariates are generated randomly following a multivariate gaussian distribution:
\begin{equation}
\left[\begin{array}{c} x_1 \\ x_2 \\ ... \\ x_{10} \end{array}\right]= \mathcal{N}\left(\left[\begin{array}{c} \mu_1 \\ \mu_2 \\ ... \\ \mu_{10} \end{array}\right], \Sigma \right)
\label{Eq2}
\end{equation}
with $\mu_1, \mu_2, ..., \mu_{10}$ the covariates means and $\Sigma$ the covariance matrix. To generate $\Sigma$ we choose 10 eigenvalues $(\lambda_1, \lambda_2, ... , \lambda_{10})$, and sample a random orthogonal matrix $O$ of size $9 \times 9$. An intermediate covariance matrix $\Sigma^{\prime}$ is obtained through:
\begin{equation}
\Sigma^\prime = O\begin{bmatrix}\lambda_{1}& 0 &...&0\\0 & \lambda_2 & \ddots & \vdots \\\vdots &\ddots & \ddots & 0 \\0&...&0&\lambda_{9}\end{bmatrix}O^t
\label{Eq3}
\end{equation}
The final covariance matrix $\Sigma$ is then generated concatenating $\Sigma^\prime$ with $\lambda_{10}$ in a block-diagonal matrix:
\begin{equation}
\Sigma=\left[
\begin{array}{c|c}

  \raisebox{-15pt}{\huge\mbox{{$\Sigma^{\prime}$}}} & 0\\[-4ex]
 & \vdots  \\[-0.5ex]
 & 0  \\ \hline
   0 \cdots 0 & \lambda_{10}
\end{array}
\right]
\label{Eq4}
\end{equation}
Generating covariates $\bm{x}=(x_1, x_2, ..., x_{10})$ according to equations (\ref{Eq2}), (\ref{Eq3}) and (\ref{Eq4}) provides a set of $9$ correlated covariates $x_1, x_2, ... ,x_9$ and an independent covariate $x_{10}$. Once covariates have been generated, we randomly generate their associated outcomes $y$ through a logistic model. We consider a situation where the logarithm of the odds is a strongly non-linear function of the covariates:
\begin{equation}
\log\frac{p\big(y=1 | x_1,..., x_{10}\big)}{p\big(y=0 | x_1,..., x_{10}\big)}= a_0 +a_1 x_1 +a_2 x_2 + a_3 x_3 + a_{4}x_1 x_2 + a_{5}x_3I(x_4>0) + a_{6}x_5^2 I(x_6>0) + a_7 x_7 I(x_8 x_9>0)
\label{Eq5}
\end{equation}
with $(a_0, a_1, ..., a_7)$ a set of constants. Covariates $x_1$, $x_2$, ..., $x_9$ are associated with the outcome $y$ contrary to $x_{10}$ that remains independent of all other variables. The strongly non-linear case given by equation (\ref{Eq5}) corresponds to a generic situation with complex interactions that cannot be accounted for by simple generalized linear models.

Each simulation consists in generating randomly $n_{gen}=10000$ records and then in adding randomly $n_{dup}=2000$ duplicates ($17\%$ of the total number of records). Each of the $n_{gen}$ original records are drawn from the probability distributions given by equations (\ref{Eq2}) and (\ref{Eq5}), and is then attributed randomly to one of the $n_h=5$ hospitals with uniform probability. To generate duplicates we randomly draw one of the $n_{gen}$ original records and one of the $n_h$ hospitals. We then add a duplicate of the drawn record to the drawn hospital unless a duplicate of the original record already exists in the hospital that has been drawn, thus ensuring that definition \ref{Def1} is fulfilled. We repeat this procedure until $n_{dup}$ duplicated records have been added to the hospitals datasets.

Unless stated otherwise we consider centered covariates $(\mu_1, \mu_2,... , \mu_  {10})=(0,\ 0, ...,\ 0)$ generated through equation (\ref{Eq2}) using eigenvalues $(\lambda_1, \lambda_2, ... , \lambda_{10})=(1.0,\ 1.2,\ 1.4,\ 1.6,\ 1.8,\ 2.0,\ 2.2,\ 2.4,\ 2.6,\ 2.8)$. Outcomes are generated using equation (\ref{Eq5}) with parameters $(a_0, a_1, a_2, a_3, a_4, a_5, a_6, a_7)= (-2,\ 0.4,\ 0.8,\ 1.2,\ 0.4,\ 1.2,\ 3.0,\ 2.0)$ leading to $\sim 47\%$ of records associated to a positive outcome $y=1$. Cross-validation is realized with $k=5$ folds. The code used for the simulations is available in the Supplemental material.

\section*{Results}

\subsection*{Model definition and dataset partitioning strategies}

\begin{figure}
\centering
\includegraphics[width=8cm]{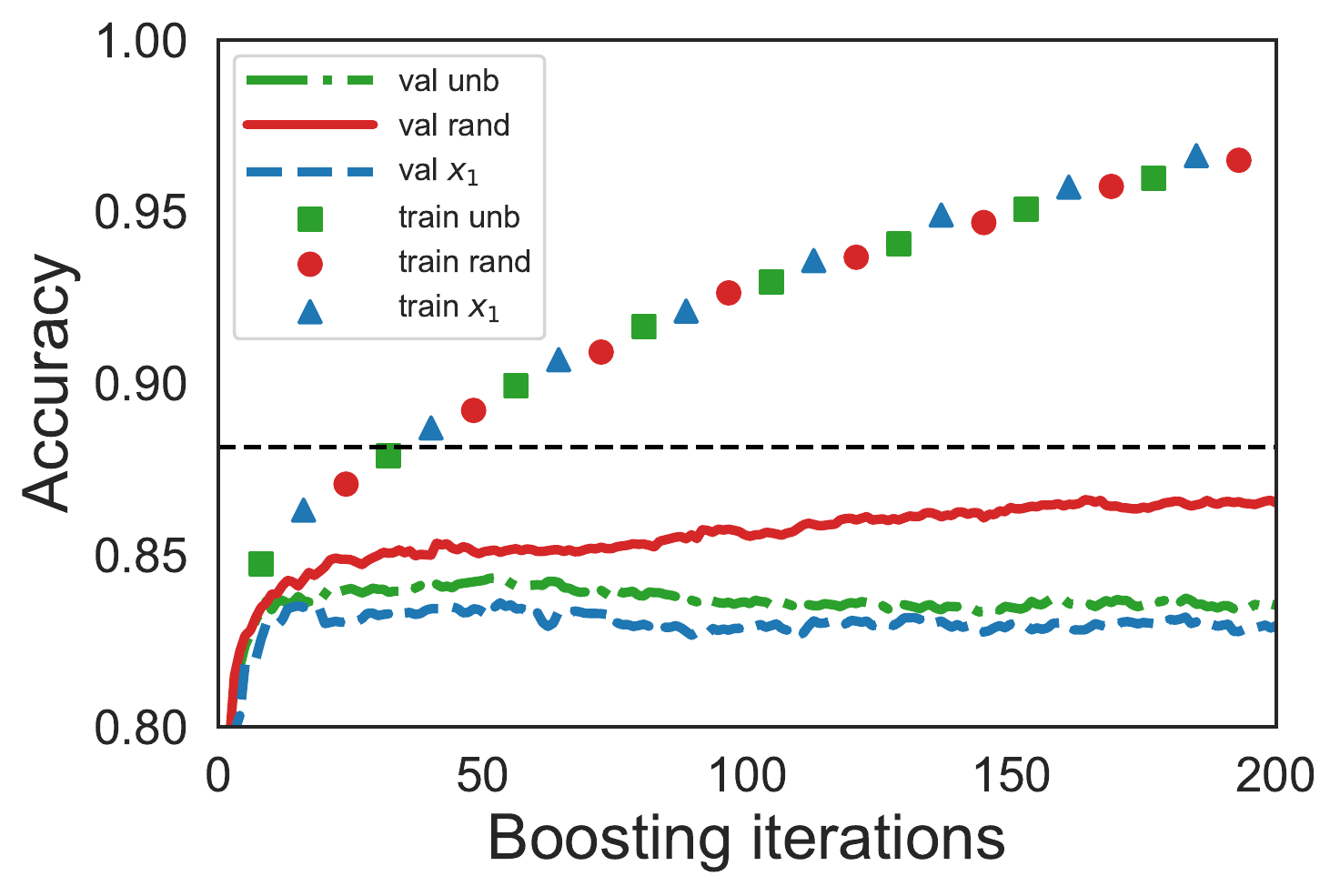}
\caption{Accuracies computed through cross-validation as a function of the number of boosting iterations. Symbols and curves correspond respectively to training accuracies and validation accuracies. Green, red and blue colors correspond respectively to \textit{unbiased}, \textit{random} and \textit{stratified along $x_1$} fold-partitioning strategies. \textit{Unbiased} validation accuracy lies between the over-optimistic \textit{random} and the pessimistic \textit{$x_1$-stratified} estimates. Horizontal black dashed line indicates the theoretical optimal accuracy $Accuracy_{opt}$.}
\label{Fig2}
\end{figure}

High-dimensional non-linear problems on tabular data are commonly modeled using gradient boosting \cite{hastie_elements_2009} and we use here its implementation in XGBoost library \cite{chen_xgboost:_2016}. We consider trees of depth 3 that are added successively during $200$ boosting iterations. The learning parameter is set to $0.6$ and we use the binary logistic loss function. Although it is currently not possible to apply directly XGBoost in a federated learning setting, protocols are being developed to circumvent this difficulty \cite{liu_boosting_2019, cheng_secureboost:_2019}. These computational considerations are not related to the data leakage issue under consideration and, for the sake of simplicity, in our simulations XGBoost is applied on physically centralized datasets simulating in this way gradient boosting in a federated learning setting.

For each simulation we first generate a dataset without duplicates and we fit and validate a gradient boosting model using the classical cross-validation methodology, measuring thus an unbiased estimate of performances. We then add duplicates and consider various validation strategies. \textit{Random partitioning} consists in partitioning randomly each hospital dataset in $k$ folds of the same size. \textit{Stratified partitioning} consists in choosing first a stratifying covariate $x_{str}$ and a set of thresholds $\{t_0, t_1,..., t_k\}$. Each record is then attributed to the fold $i$ that fulfills $t_{i}<x_{str}<t_{i+1}$. We choose thresholds $t_i$ in such a way that the number of records in each global fold $\mathcal{D}_i$ is the same. Once fold-partitioning is realized, gradient boosting models are fitted and validated on these folds.

\subsection*{Model training and validation}

For each $i\in 1,..,k$ the model is trained on all the folds apart from fold $\mathcal{D}_i$ and its training performances are measured on the same folds. Figure \ref{Fig2} shows the training learning curves obtained during boosting for \textit{unbiased} (squares), \textit{random} (circles) and \textit{stratified along $x_1$} (triangles) strategies. Training accuracies increase monotonously as the model learns from the training dataset, and the learning speed does not depend on the fold-partitioning strategy that is adopted. Indeed, training accuracies do not depend on strategy-dependent data leakage between training and validation folds. 

For each $i\in 1,..,k$ the model trained on folds $\mathcal{D}_j$ with $j\neq i$ is validated on fold $\mathcal{D}_i$. Figure \ref{Fig2} shows the variation of validation accuracies during training (solid and dashed lines). When a deduplicated dataset is used (green dashed curve), the unbiased validation accuracy increases during the first 50 boosting iterations and then saturates at a value that is lower than the optimal accuracy that a predictive model could reach (straight dashed black line). As expected the validation accuracy remains lower than the optimal accuracy. When duplicates are added and \textit{random} fold-partitioning strategy is used, the estimated validation accuracy (solid red curve) is biased by data leakage and it increases monotonously until falsely reaching a high accuracy. When \textit{stratified along $x_{1}$} strategy is adopted, duplicates of a given record are grouped in the same fold and definition \ref{Def3} is fulfilled: there is consequently no over-optimistic bias and the estimated accuracy remains close to the unbiased one, but a small pessimistic bias is observed that is due to inter-fold heterogeneity. Applying \textit{random} fold-partitioning strategy on a dataset with duplicates therefore misses the saturation of the performances observed in \textit{unbiased} and \textit{stratified} cases.

\begin{figure}
\centering
\includegraphics[width=8cm]{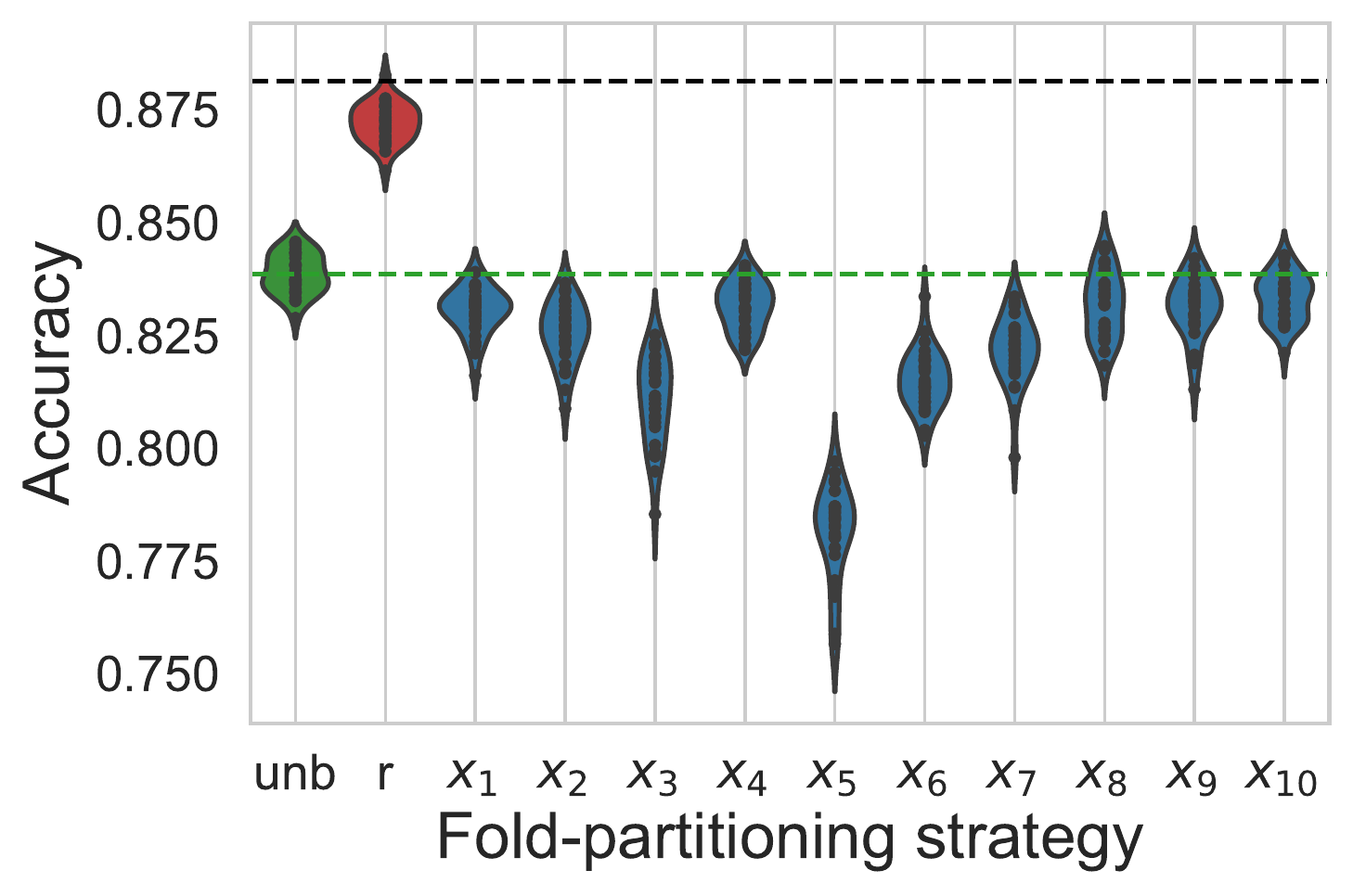}
\caption{Violin plots for cross-validation estimates of accuracy adopting either an \textit{unbiased} (green), a \textit{random} (red) or a $x_1, x_2, ..., x_{10}$\textit{-stratified} (blue) fold-partitioning strategy and running 30 simulations. Horizontal black and green dashed lines correspond respectively to the optimal accuracy that a model could reach $Accuracy_{opt}$ and to the mean unbiased estimate of the accuracy $Accuracy_{unb}$ reached by the model under consideration. Whereas \textit{random} fold-partitioning leads to over-optimistic estimates of accuracy, $x_1, x_2, ..., x_{10}$\textit{-stratified} estimates feature pessimistic biases of various sizes.}
\label{Fig3}
\end{figure}

\subsection*{Bias and feature importance}

We ran additional simulations to better understand the implications of the choice of a stratifying covariate. Figure \ref{Fig3} shows validation accuracies obtained after 200 boosting iterations for 30 simulations using the same set of generating parameters $\Sigma$ and $(a_0, a_1, ..., a_7)$ but applying various fold-partitioning strategies. The distribution of estimated accuracies are shown as violin plots. Whereas \textit{random} fold-partitioning always leads to over-optimistic estimates of accuracy (red) compared to the \textit{unbiased} estimates obtained without duplicates (green), other stratification strategies lead to accuracy estimates that feature pessimistic biases of variable importance (blue). Whereas $x_5$ stratification leads to a pessimistic bias of roughly $5\%$, stratifying along $x_1$, $x_2$, $x_4$, $x_8$, $x_9$ or $x_{10}$ leads to estimates that are close to the unbiased ones. The arbitrary choice of a stratifying covariate may consequently lead to important pessimistic biases. The closeness with the unbiased estimates of the estimates obtained having stratified along $x_{10}$ was predictable as $x_{10}$ is an independent variable corresponding thus to an ideal stratifying covariate that does not induce inter-fold heterogeneity. 

\begin{figure}
\centering
\includegraphics[width=8cm]{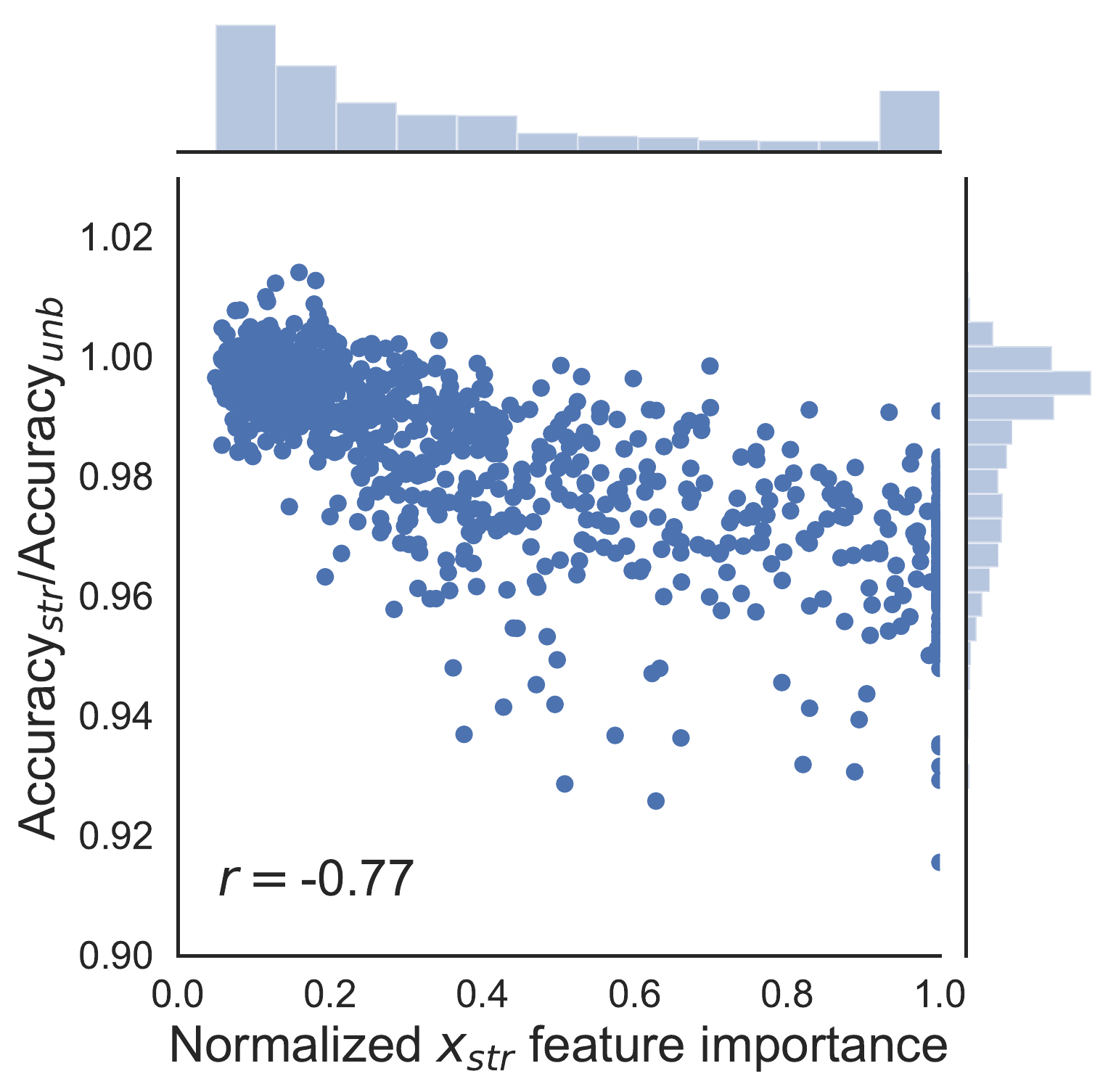}
\caption{Ratio of \textit{$x_{str}$-stratified} estimate of accuracy over the \textit{unbiased} estimate of accuracy plotted with respect to the normalized importance of the stratifying covariate $x_{str}$ (\textit{see text}). 100 datasets are generated corresponding to different $\{\Sigma,\bm{a}\}$ and, for each dataset, each covariate is taken successively as stratifying covariate. A Pearson correlation coefficient $r=-0.77$ is measured.}
\label{Fig4}
\end{figure}

To better characterize pessimistic biases we run additional simulations varying also the covariate covariance matrix $\Sigma$ and the outcome generating parameters $\bm{a}=(a_0, a_1, ..., a_7)$. To generate each new covariance matrix $\Sigma$ we draw random eigenvalues $\lambda_1, \lambda_2, ...,\lambda_{10}$ from a uniform distribution on the interval $[1,3]$, we generate a new random orthonormal matrix $O$ and we combine them through equations (\ref{Eq3}) and \ref{Eq4}. Each new outcome generating parameter $a_i$, $i=0,..,6$ is drawn from a uniform distribution on $[-5,5]$. We draw 100 different sets of parameters $\{\Sigma, \bm{a}\}$ and, for each set of generating parameters, we generate a dataset. For each dataset, we fit and validate a gradient boosting model using classical cross-validation and measuring its unbiased validation accuracy $Accuracy_{unb}$. We then add duplicates and fit successively 10 gradient boosting models corresponding each to \textit{stratified cross-validation} having stratified along one of the 10 covariates. Figure \ref{Fig4} shows for each one of the 1000 models the validation accuracy $Accuracy_{str}$  measured having stratified along $x_{str}$ divided by the unbiased estimate $Accuracy_{unb}$ obtained on the same dataset, plotted with respect to normalized feature importance of $x_{str}$ measured during the training of the unbiased model (cf. \cite{hastie_elements_2009} for feature importance computation). We observe a negative Pearson correlation coefficient of $-0.77$ between the pessimistic bias on accuracy and the normalized importance of the stratifying covariate: the stronger the importance of the stratifying covariate, the bigger the pessimistic bias.

\section*{Discussion}

The data leakage phenomenon at play in our simulations is specific neither to the XGBoost model nor to the synthetic data under scrutiny, and the risk of duplicate-caused data leakage should be addressed whatever the scientific problem at stake. \textit{Stratified cross-validation} methodology avoids data leakage between folds by fulfilling inter-fold deduplication (definition \ref{Def3}) without requiring full deduplication (definitions \ref{Def1} and \ref{Def2}), providing thus a validation methodology that is robust to the presence of undetected duplicates (Figures \ref{Fig2} and \ref{Fig3}). Although \textit{stratified cross-validation} avoids an over-optimistic bias due to data leakage between folds, it may be subject to a pessimistic bias due to inter-folds heterogeneity (Figures \ref{Fig2} and \ref{Fig3}). In order to limit this undesired pessimistic bias it appears optimal to choose a stratifying covariate that is weakly associated to the other covariates and to the outcome (Figure \ref{Fig4}). Determining \textit{a priori} which covariate is weakly or strongly associated to the other covariates and to the outcome is challenging, and can rely either on prior knowledge of the problem under scrutiny or on preliminary fitting of the model on a local dataset. In the special case of the date of birth or a personal identifier being available, hashing it provides an ideal stratifying covariate that is shared by duplicates. Another difficulty might arise when records related to a given individual are not perfectly equal as it is the case when duplicated records correspond to different recording events such as hospital admissions. But perfect equality of duplicated records is not necessary to apply \textit{stratified cross-validation}: identifying a single stratifying covariate, the value of which is shared among records related to a given individual, is indeed sufficient. 

\textit{Stratified cross-validation} methodology presents generic limitations. Firstly when duplicated records are so different that no shared covariate can be found to stratify, it appears impossible to ensure through stratification that two records related to a given individual are present in the same fold. Secondly when data curators are not hospitals but directly individuals, such as for instance in the case of data stored in mobile phones \cite{bonawitz_towards_2019}, it is impossible to partition datasets in folds as each dataset corresponds to a single record. Thirdly pseudonymization is often applied to records prior to their analysis (\textit{e.g.} date shifting, quantization, noise addition \cite{emam_anonymizing_2013, dwork_algorithmic_2014}). Applying hospital-specific or probabilistic pseudonymization may break the equality of stratifying covariates among duplicates. Fourthly a subpopulation may be more subject to data duplication, thus inducing its over-representation in the validation fold. We moreover underline that the presented simulations are mostly illustrative: the size of biases measured in this article depend strongly on the data generation procedure, the model and the metric at stake. 

\section*{Conclusion}

When a model is trained and validated in a privacy-preserving federated learning setting, the presence of duplicated records may lead to over-optimistically biased estimates of its performances. We have shown that \textit{stratified cross-validation} methodology can be used to avoid this over-optimistic bias without fully deduplicating records, although at the possible cost of a pessimistic bias that can be minimized by carefully choosing the stratifying covariate. We underline that \textit{stratified cross-validation}, although of special importance in the case of federated learning where full deduplication is often unfeasible, also applies to the case of a centralized dataset with undetected duplicates and can therefore be used as an easy-to-implement sanity check. Although of possibly broad application, \textit{stratified cross-validation} is only a partial solution to duplicate problems: full deduplication remains optimal to ensure databases integrity.

\section*{Acknowledgements}

We thank the partners of "Healthchain" consortium for fruitfull discussions and E. Diard for her help with figure design.

\section*{Funding}

This project is supported  by Bpifrance as part of the "Healthchain" project, which resulted from the "Digital Investments Program for the major challenges of the future" RFP. As part of the "Healthchain" project, a consortium coordinated by Owkin (a private company) has been established, including the Substra association, Apricity (a private company), the Assistance Publique des Hôpitaux de Paris, the University Hospital Center of Nantes, the Léon Bérard Center, the French National Center for Scientific Research, the Ecole Polytechnique, the Institut Curie and the University of Paris.

\section*{Contributions}
R.B. designed the methodology, conducted the simulations and wrote the manuscript.

R.G. and M.B. provided technical advice and manuscript feedbacks.

R.P. oversaw the project and helped with manuscript writing.

\section*{Conflict of interest statement}
None.

\section*{Supplemental material}

\subsection*{Code availability}

The Jupyter notebook containing all the simulation code is available at https://doi.org/10.5281/zenodo.3614900 under Apache2.0 license.

\subsection*{Computation of the optimal accuracy}

For each set of covariates $(x_1$, $x_2$, ..., $x_{10})$, the exact probability $p_1$ of a positive outcome $y=1$ can be computed using equation (\ref{Eq5}). An ideal predictive model that optimizes its accuracy would predict $\tilde{y}=1$ when $p_1>0.5$ and $\tilde{y}=0$ otherwise. The average accuracy of such a model computed over the population of interest writes (see equations (\ref{Eq2}), (\ref{Eq3}), (\ref{Eq4})):
\begin{equation}
Accuracy_{opt} = \int \Big[p_1I(p_1>0.5) + (1-p_1)I(p_1<0.5)\Big] \mathcal{N}\Big(\bm{\mu} , \Sigma \Big) dx_1dx_2...dx_{10}
\end{equation}
We estimate $Accuracy_{opt}$ through a Monte Carlo computation. We draw randomly 100000 records using the multivariate Gaussian $\mathcal{N}(\bm{\mu} , \Sigma)$ and we average the value of $p_1I(p_1>0.5) + (1-p_1)I(p_1<0.5)$ over the records. We measure an optimal accuracy of $Accuracy_{opt}=0.88$.

\subsection*{Data visualization}

\begin{figure}
\centering
\includegraphics[width=14cm]{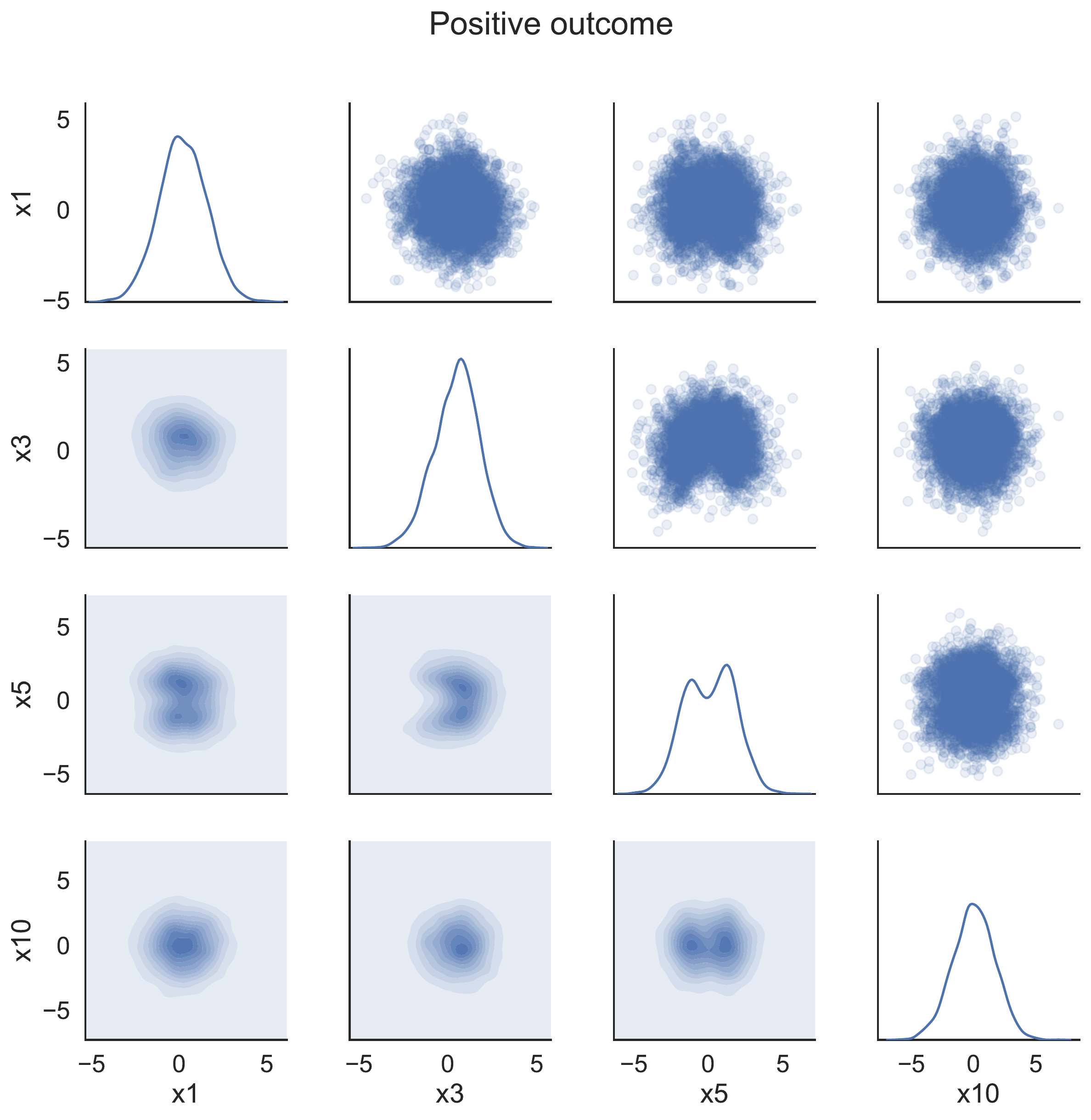}
\caption{Diagonal: distributions of positive outcomes relatively to a single covariate. Upper-edge and lower-edge: scatter plots and density plots of positive outcomes relatively to two covariates.}
\label{FigSI_1}
\end{figure}

In our simulations, records are generated following equations (\ref{Eq2}), (\ref{Eq3}), (\ref{Eq4}) and (\ref{Eq5}). Covariates are drawn from a multivariate Gaussian distribution (equation (\ref{Eq2})) and outcomes are drawn randomly from the covariates using a logistic function (equation (\ref{Eq5})). Figure \ref{FigSI_1} represents the distribution of positive outcomes relatively to covariates $x_1$, $x_3$, $x_5$, $x_{10}$ obtained having drawn randomly 10000 records. These 4 covariates have been chosen arbitrarily in order to illustrate the structure of the datasets under scrutiny.

\bibliography{Biblio} 

\begin{thebibliography}{10}

\bibitem{esteva_dermatologist-level_2017}
Andre Esteva, Brett Kuprel, Roberto~A. Novoa, Justin Ko, Susan~M. Swetter,
  Helen~M. Blau, and Sebastian Thrun.
\newblock Dermatologist-level classification of skin cancer with deep neural
  networks.
\newblock {\em Nature}, 542(7639):115--118, 2017.

\bibitem{hosny_artificial_2018}
Ahmed Hosny, Chintan Parmar, John Quackenbush, Lawrence~H. Schwartz, and Hugo
  J. W.~L. Aerts.
\newblock Artificial intelligence in radiology.
\newblock {\em Nat. Rev. Cancer}, 18(8):500--510, 2018.

\bibitem{komorowski_artificial_2018}
Matthieu Komorowski, Leo~A. Celi, Omar Badawi, Anthony~C. Gordon, and A.~Aldo
  Faisal.
\newblock The {Artificial} {Intelligence} {Clinician} learns optimal treatment
  strategies for sepsis in intensive care.
\newblock {\em Nat. Med.}, 24(11):1716--1720, 2018.

\bibitem{rajkomar_scalable_2018}
Alvin Rajkomar, Eyal Oren, Kai Chen, Andrew~M. Dai, Nissan Hajaj, Michaela
  Hardt, Peter~J. Liu, Xiaobing Liu, Jake Marcus, Mimi Sun, Patrik Sundberg,
  Hector Yee, Kun Zhang, Yi~Zhang, Gerardo Flores, Gavin~E. Duggan, Jamie
  Irvine, Quoc Le, Kurt Litsch, Alexander Mossin, Justin Tansuwan, De~Wang,
  James Wexler, Jimbo Wilson, Dana Ludwig, Samuel~L. Volchenboum, Katherine
  Chou, Michael Pearson, Srinivasan Madabushi, Nigam~H. Shah, Atul~J. Butte,
  Michael~D. Howell, Claire Cui, Greg~S. Corrado, and Jeffrey Dean.
\newblock Scalable and accurate deep learning with electronic health records.
\newblock {\em NPJ Digit Med}, 1:18, 2018.

\bibitem{rahimian_predicting_2018}
Fatemeh Rahimian, Gholamreza Salimi-Khorshidi, Amir~H. Payberah, Jenny Tran,
  Roberto Ayala~Solares, Francesca Raimondi, Milad Nazarzadeh, Dexter Canoy,
  and Kazem Rahimi.
\newblock Predicting the risk of emergency admission with machine learning:
  {Development} and validation using linked electronic health records.
\newblock {\em PLoS Med.}, 15(11):e1002695, 2018.

\bibitem{hastie_elements_2009}
Trevor Hastie, Robert Tibshirani, and Jerome Friedman.
\newblock {\em The {Elements} of {Statistical} {Learning}: {Data} {Mining},
  {Inference}, and {Prediction}, {Second} {Edition}}.
\newblock Springer {Series} in {Statistics}. Springer-Verlag, New York, 2
  edition, 2009.

\bibitem{van_der_ploeg_modern_2014}
Tjeerd van~der Ploeg, Peter~C. Austin, and Ewout~W. Steyerberg.
\newblock Modern modelling techniques are data hungry: a simulation study for
  predicting dichotomous endpoints.
\newblock {\em BMC Med Res Methodol}, 14:137, December 2014.

\bibitem{powles_google_2017}
Julia Powles and Hal Hodson.
\newblock Google {DeepMind} and healthcare in an age of algorithms.
\newblock {\em Health Technol (Berl)}, 7(4):351--367, 2017.

\bibitem{caldicott_review_2016}
Fiona Caldicott.
\newblock Review of data security, consent and opt-outs, 2016.

\bibitem{homer_resolving_2008}
Nils Homer, Szabolcs Szelinger, Margot Redman, David Duggan, Waibhav Tembe,
  Jill Muehling, John~V. Pearson, Dietrich~A. Stephan, Stanley~F. Nelson, and
  David~W. Craig.
\newblock Resolving individuals contributing trace amounts of {DNA} to highly
  complex mixtures using high-density {SNP} genotyping microarrays.
\newblock {\em PLoS Genet.}, 4(8):e1000167, August 2008.

\bibitem{bohannon_genetics._2013}
John Bohannon.
\newblock Genetics. {Genealogy} databases enable naming of anonymous {DNA}
  donors.
\newblock {\em Science}, 339(6117):262, January 2013.

\bibitem{gymrek_identifying_2013}
Melissa Gymrek, Amy~L. McGuire, David Golan, Eran Halperin, and Yaniv Erlich.
\newblock Identifying personal genomes by surname inference.
\newblock {\em Science}, 339(6117):321--324, January 2013.

\bibitem{rocher_estimating_2019}
Luc Rocher, Julien~M. Hendrickx, and Yves-Alexandre de~Montjoye.
\newblock Estimating the success of re-identifications in incomplete datasets
  using generative models.
\newblock {\em Nat Commun}, 10(1):3069, 2019.

\bibitem{price_privacy_2019}
W.~Nicholson Price and I.~Glenn Cohen.
\newblock Privacy in the age of medical big data.
\newblock {\em Nat. Med.}, 25(1):37--43, 2019.

\bibitem{aggarwal_k-anonymity_2005}
Charu~C. Aggarwal.
\newblock On {K}-anonymity and the {Curse} of {Dimensionality}.
\newblock In {\em Proceedings of the 31st {International} {Conference} on
  {Very} {Large} {Data} {Bases}}, {VLDB} '05, pages 901--909. VLDB Endowment,
  2005.
\newblock event-place: Trondheim, Norway.

\bibitem{brickell_cost_2008}
Justin Brickell and Vitaly Shmatikov.
\newblock The {Cost} of {Privacy}: {Destruction} of {Data}-mining {Utility} in
  {Anonymized} {Data} {Publishing}.
\newblock In {\em Proceedings of the 14th {ACM} {SIGKDD} {International}
  {Conference} on {Knowledge} {Discovery} and {Data} {Mining}}, {KDD} '08,
  pages 70--78, New York, NY, USA, 2008. ACM.
\newblock event-place: Las Vegas, Nevada, USA.

\bibitem{de_montjoye_privacy-conscientious_2018}
Yves-Alexandre de~Montjoye, Sébastien Gambs, Vincent Blondel, Geoffrey
  Canright, Nicolas de~Cordes, Sébastien Deletaille, Kenth Engø-Monsen,
  Manuel Garcia-Herranz, Jake Kendall, Cameron Kerry, Gautier Krings, Emmanuel
  Letouzé, Miguel Luengo-Oroz, Nuria Oliver, Luc Rocher, Alex Rutherford,
  Zbigniew Smoreda, Jessica Steele, Erik Wetter, Alex~Sandy Pentland, and Linus
  Bengtsson.
\newblock On the privacy-conscientious use of mobile phone data.
\newblock {\em Sci Data}, 5:180286, 2018.

\bibitem{vest_hospitals_2018}
Joshua~R. Vest and Kosali Simon.
\newblock Hospitals' adoption of intra-system information exchange is
  negatively associated with inter-system information exchange.
\newblock {\em J Am Med Inform Assoc}, 25(9):1189--1196, 2018.

\bibitem{wu_grid_2012}
Yuan Wu, Xiaoqian Jiang, Jihoon Kim, and Lucila Ohno-Machado.
\newblock Grid {Binary} {LOgistic} {REgression} ({GLORE}): building shared
  models without sharing data.
\newblock {\em J Am Med Inform Assoc}, 19(5):758--764, October 2012.

\bibitem{lu_webdisco:_2015}
Chia-Lun Lu, Shuang Wang, Zhanglong Ji, Yuan Wu, Li~Xiong, Xiaoqian Jiang, and
  Lucila Ohno-Machado.
\newblock {WebDISCO}: a web service for distributed cox model learning without
  patient-level data sharing.
\newblock {\em J Am Med Inform Assoc}, 22(6):1212--1219, November 2015.

\bibitem{shokri_privacy-preserving_2015}
Reza Shokri and Vitaly Shmatikov.
\newblock Privacy-{Preserving} {Deep} {Learning}.
\newblock In {\em Proceedings of the 22Nd {ACM} {SIGSAC} {Conference} on
  {Computer} and {Communications} {Security}}, {CCS} '15, pages 1310--1321, New
  York, NY, USA, 2015. ACM.
\newblock event-place: Denver, Colorado, USA.

\bibitem{mcmahan_communication-efficient_2017}
H.~Brendan McMahan, Eider Moore, Daniel Ramage, Seth Hampson, and Blaise
  Agüera~y Arcas.
\newblock Communication-{Efficient} {Learning} of {Deep} {Networks} from
  {Decentralized} {Data}.
\newblock {\em arXiv:1602.05629 [cs]}, February 2017.
\newblock arXiv: 1602.05629.

\bibitem{bonawitz_practical_2017}
Keith Bonawitz, Vladimir Ivanov, Ben Kreuter, Antonio Marcedone, H.~Brendan
  McMahan, Sarvar Patel, Daniel Ramage, Aaron Segal, and Karn Seth.
\newblock Practical {Secure} {Aggregation} for {Privacy}-{Preserving} {Machine}
  {Learning}.
\newblock In {\em Proceedings of the 2017 {ACM} {SIGSAC} {Conference} on
  {Computer} and {Communications} {Security}}, {CCS} '17, pages 1175--1191, New
  York, NY, USA, 2017. ACM.
\newblock event-place: Dallas, Texas, USA.

\bibitem{kairouz_advances_2019}
Peter Kairouz, H.~Brendan McMahan, Brendan Avent, Aurélien Bellet, Mehdi
  Bennis, Arjun~Nitin Bhagoji, Keith Bonawitz, Zachary Charles, Graham Cormode,
  Rachel Cummings, Rafael G.~L. D'Oliveira, Salim~El Rouayheb, David Evans,
  Josh Gardner, Zachary Garrett, Adrià Gascón, Badih Ghazi, Phillip~B.
  Gibbons, Marco Gruteser, Zaid Harchaoui, Chaoyang He, Lie He, Zhouyuan Huo,
  Ben Hutchinson, Justin Hsu, Martin Jaggi, Tara Javidi, Gauri Joshi, Mikhail
  Khodak, Jakub Konečný, Aleksandra Korolova, Farinaz Koushanfar, Sanmi
  Koyejo, Tancrède Lepoint, Yang Liu, Prateek Mittal, Mehryar Mohri, Richard
  Nock, Ayfer Özgür, Rasmus Pagh, Mariana Raykova, Hang Qi, Daniel Ramage,
  Ramesh Raskar, Dawn Song, Weikang Song, Sebastian~U. Stich, Ziteng Sun,
  Ananda~Theertha Suresh, Florian Tramèr, Praneeth Vepakomma, Jianyu Wang,
  Li~Xiong, Zheng Xu, Qiang Yang, Felix~X. Yu, Han Yu, and Sen Zhao.
\newblock Advances and {Open} {Problems} in {Federated} {Learning}.
\newblock {\em arXiv:1912.04977 [cs, stat]}, December 2019.
\newblock arXiv: 1912.04977.

\bibitem{bonawitz_towards_2019}
Keith Bonawitz, Hubert Eichner, Wolfgang Grieskamp, Dzmitry Huba, Alex
  Ingerman, Vladimir Ivanov, Chloe Kiddon, Jakub Konečný, Stefano Mazzocchi,
  H.~Brendan McMahan, Timon Van~Overveldt, David Petrou, Daniel Ramage, and
  Jason Roselander.
\newblock Towards {Federated} {Learning} at {Scale}: {System} {Design}.
\newblock {\em arXiv:1902.01046 [cs, stat]}, March 2019.
\newblock arXiv: 1902.01046.

\bibitem{raisaro_addressing_2017}
Jean~Louis Raisaro, Florian Tramèr, Zhanglong Ji, Diyue Bu, Yongan Zhao, Knox
  Carey, David Lloyd, Heidi Sofia, Dixie Baker, Paul Flicek, Suyash
  Shringarpure, Carlos Bustamante, Shuang Wang, Xiaoqian Jiang, Lucila
  Ohno-Machado, Haixu Tang, XiaoFeng Wang, and Jean-Pierre Hubaux.
\newblock Addressing {Beacon} re-identification attacks: quantification and
  mitigation of privacy risks.
\newblock {\em J Am Med Inform Assoc}, 24(4):799--805, July 2017.

\bibitem{raisaro_medco:_2019}
Jean~Louis Raisaro, Juan~Ramon Troncoso-Pastoriza, Mickael Misbach, Joao~Sa
  Sousa, Sylvain Pradervand, Edoardo Missiaglia, Olivier Michielin, Bryan Ford,
  and Jean-Pierre Hubaux.
\newblock {MedCo}: {Enabling} {Secure} and {Privacy}-{Preserving} {Exploration}
  of {Distributed} {Clinical} and {Genomic} {Data}.
\newblock {\em IEEE/ACM Trans Comput Biol Bioinform}, 16(4):1328--1341, August
  2019.

\bibitem{ryffel_generic_2018}
Theo Ryffel, Andrew Trask, Morten Dahl, Bobby Wagner, Jason Mancuso, Daniel
  Rueckert, and Jonathan Passerat-Palmbach.
\newblock A generic framework for privacy preserving deep learning.
\newblock {\em arXiv:1811.04017 [cs, stat]}, November 2018.
\newblock arXiv: 1811.04017.

\bibitem{galtier_substra:_2019}
Mathieu~N. Galtier and Camille Marini.
\newblock Substra: a framework for privacy-preserving, traceable and
  collaborative {Machine} {Learning}.
\newblock {\em arXiv:1910.11567 [cs]}, October 2019.
\newblock arXiv: 1910.11567.

\bibitem{duan_learning_2019}
Rui Duan, Mary~Regina Boland, Zixuan Liu, Yue Liu, Howard~H. Chang, Hua Xu,
  Haitao Chu, Christopher~H. Schmid, Christopher~B. Forrest, John~H. Holmes,
  Martijn~J. Schuemie, Jesse~A. Berlin, Jason~H. Moore, and Yong Chen.
\newblock Learning from electronic health records across multiple sites: {A}
  communication-efficient and privacy-preserving distributed algorithm.
\newblock {\em J Am Med Inform Assoc}, December 2019.

\bibitem{lazer_big_2014}
David Lazer, Ryan Kennedy, Gary King, and Alessandro Vespignani.
\newblock Big data. {The} parable of {Google} {Flu}: traps in big data
  analysis.
\newblock {\em Science}, 343(6176):1203--1205, March 2014.

\bibitem{dressel_accuracy_2018}
Julia Dressel and Hany Farid.
\newblock The accuracy, fairness, and limits of predicting recidivism.
\newblock {\em Sci Adv}, 4(1):eaao5580, 2018.

\bibitem{kiraly_nips_2018}
Franz~J. Király, Bilal Mateen, and Raphael Sonabend.
\newblock {NIPS} - {Not} {Even} {Wrong}? {A} {Systematic} {Review} of
  {Empirically} {Complete} {Demonstrations} of {Algorithmic} {Effectiveness} in
  the {Machine} {Learning} and {Artificial} {Intelligence} {Literature}.
\newblock {\em arXiv:1812.07519 [cs, stat]}, December 2018.
\newblock arXiv: 1812.07519.

\bibitem{park_methodologic_2018}
Seong~Ho Park and Kyunghwa Han.
\newblock Methodologic {Guide} for {Evaluating} {Clinical} {Performance} and
  {Effect} of {Artificial} {Intelligence} {Technology} for {Medical}
  {Diagnosis} and {Prediction}.
\newblock {\em Radiology}, 286(3):800--809, 2018.

\bibitem{vollmer_machine_2018}
Sebastian Vollmer, Bilal~A. Mateen, Gergo Bohner, Franz~J. Király, Rayid
  Ghani, Pall Jonsson, Sarah Cumbers, Adrian Jonas, Katherine S.~L. McAllister,
  Puja Myles, David Granger, Mark Birse, Richard Branson, Karel~GM Moons,
  Gary~S. Collins, John P.~A. Ioannidis, Chris Holmes, and Harry Hemingway.
\newblock Machine learning and {AI} research for {Patient} {Benefit}: 20
  {Critical} {Questions} on {Transparency}, {Replicability}, {Ethics} and
  {Effectiveness}.
\newblock {\em arXiv:1812.10404 [cs, stat]}, December 2018.
\newblock arXiv: 1812.10404.

\bibitem{kaufman_leakage_2012}
Shachar Kaufman, Saharon Rosset, Claudia Perlich, and Ori Stitelman.
\newblock Leakage in {Data} {Mining}: {Formulation}, {Detection}, and
  {Avoidance}.
\newblock {\em ACM Trans. Knowl. Discov. Data}, 6(4):15:1--15:21, December
  2012.

\bibitem{harron_evaluating_2014}
Katie Harron, Angie Wade, Ruth Gilbert, Berit Muller-Pebody, and Harvey
  Goldstein.
\newblock Evaluating bias due to data linkage error in electronic healthcare
  records.
\newblock {\em BMC Med Res Methodol}, 14:36, March 2014.

\bibitem{luo_guidelines_2016}
Wei Luo, Dinh Phung, Truyen Tran, Sunil Gupta, Santu Rana, Chandan Karmakar,
  Alistair Shilton, John Yearwood, Nevenka Dimitrova, Tu~Bao Ho, Svetha
  Venkatesh, and Michael Berk.
\newblock Guidelines for {Developing} and {Reporting} {Machine} {Learning}
  {Predictive} {Models} in {Biomedical} {Research}: {A} {Multidisciplinary}
  {View}.
\newblock {\em J. Med. Internet Res.}, 18(12):e323, 2016.

\bibitem{saeb_need_2017}
Sohrab Saeb, Luca Lonini, Arun Jayaraman, David~C. Mohr, and Konrad~P. Kording.
\newblock The need to approximate the use-case in clinical machine learning.
\newblock {\em Gigascience}, 6(5):1--9, 2017.

\bibitem{mccoy_matching_2013}
Allison~B. McCoy, Adam Wright, Michael~G. Kahn, Jason~S. Shapiro, Elmer~Victor
  Bernstam, and Dean~F. Sittig.
\newblock Matching identifiers in electronic health records: implications for
  duplicate records and patient safety.
\newblock {\em BMJ Qual Saf}, 22(3):219--224, March 2013.

\bibitem{everson_gaps_2018}
Jordan Everson and Julia Adler-Milstein.
\newblock Gaps in health information exchange between hospitals that treat many
  shared patients.
\newblock {\em J Am Med Inform Assoc}, 25(9):1114--1121, 2018.

\bibitem{harron_methodological_2015}
Katie Harron, Harvey Goldstein, and Chris Dibben.
\newblock {\em Methodological {Developments} in {Data} {Linkage}}.
\newblock John Wiley \& Sons Inc., United States, 2015.

\bibitem{vatsalan_taxonomy_2013}
Dinusha Vatsalan, Peter Christen, and Vassilios~S. Verykios.
\newblock A taxonomy of privacy-preserving record linkage techniques.
\newblock {\em Inform Syst}, 38(6):946--969, September 2013.

\bibitem{yigzaw_secure_2017}
Kassaye~Yitbarek Yigzaw, Antonis Michalas, and Johan~Gustav Bellika.
\newblock Secure and scalable deduplication of horizontally partitioned health
  data for privacy-preserving distributed statistical computation.
\newblock {\em BMC Med Inform Decis Mak}, 17(1):1, 2017.

\bibitem{laud_privacy-preserving_2018}
Peeter Laud and Alisa Pankova.
\newblock Privacy-preserving record linkage in large databases using secure
  multiparty computation.
\newblock {\em BMC Med Genomics}, 11(Suppl 4):84, October 2018.

\bibitem{diamantidis_unsupervised_2000}
N.~A. Diamantidis, D.~Karlis, and E.~A. Giakoumakis.
\newblock Unsupervised stratification of cross-validation for accuracy
  estimation.
\newblock {\em Art. Int.}, 116(1):1--16, January 2000.

\bibitem{chen_xgboost:_2016}
Tianqi Chen and Carlos Guestrin.
\newblock {XGBoost}: {A} {Scalable} {Tree} {Boosting} {System}.
\newblock {\em Proceedings of the 22nd ACM SIGKDD International Conference on
  Knowledge Discovery and Data Mining - KDD '16}, pages 785--794, 2016.
\newblock arXiv: 1603.02754.

\bibitem{liu_boosting_2019}
Yang Liu, Zhuo Ma, Ximeng Liu, Siqi Ma, Surya Nepal, and Robert Deng.
\newblock Boosting {Privately}: {Privacy}-{Preserving} {Federated} {Extreme}
  {Boosting} for {Mobile} {Crowdsensing}.
\newblock {\em arXiv:1907.10218 [cs]}, July 2019.
\newblock arXiv: 1907.10218.

\bibitem{cheng_secureboost:_2019}
Kewei Cheng, Tao Fan, Yilun Jin, Yang Liu, Tianjian Chen, and Qiang Yang.
\newblock {SecureBoost}: {A} {Lossless} {Federated} {Learning} {Framework}.
\newblock {\em arXiv:1901.08755 [cs, stat]}, January 2019.
\newblock arXiv: 1901.08755.

\bibitem{emam_anonymizing_2013}
Khaled~El Emam and Luk Arbuckle.
\newblock {\em Anonymizing {Health} {Data}: {Case} {Studies} and {Methods} to
  {Get} {You} {Started}}.
\newblock O'Reilly Media, Inc., December 2013.
\newblock Google-Books-ID: 3RtRAgAAQBAJ.

\bibitem{dwork_algorithmic_2014}
Cynthia Dwork and Aaron Roth.
\newblock The {Algorithmic} {Foundations} of {Differential} {Privacy}.
\newblock {\em Found. Trends Theor. Comput. Sci.}, 9(3–4):211--407, August
  2014.

\end{thebibliography}

\bibliographystyle{unsrt}  

\end{document}